\documentclass[11pt,a4paper]{article}
\usepackage[hyperref]{acl2020}

\usepackage{amsmath}
\usepackage{tikz}
\usepackage{mathdots}
\usepackage{yhmath}
\usepackage{cancel}
\usepackage{color}
\usepackage{siunitx}
\usepackage{array}
\usepackage{multirow}
\usepackage{amssymb}
\usepackage{gensymb}
\usepackage{tabularx}
\usetikzlibrary{fadings}
\usetikzlibrary{patterns}

\usepackage{hyphenat}
\usepackage{times}
\usepackage{latexsym}
\usepackage{cleveref}
\usepackage{url}
\usepackage{booktabs}
\usepackage{array,multirow,graphicx}
\usepackage{listings}
\usepackage{float}
\usepackage{url}
\usepackage{textcomp}

\usepackage{pifont}
\newcommand{\xmark}{\ding{55}}%

\usepackage{gb4e}

\noautomath

\def\form#1{\textit{#1}}
\def\morph#1{\textsf{\small #1}}

\usepackage{microtype}

\aclfinalcopy 

\title{Bootstrapping Techniques for Polysynthetic Morphological Analysis}

\author{William Lane \\
  Charles Darwin University \\
  {\tt william.lane@cdu.edu.au} \\\And
  Steven Bird \\
  Charles Darwin University \\
  {\tt steven.bird@cdu.edu.au} \\}

\date{}

\begin{document}
\maketitle
\begin{abstract}
    Polysynthetic languages have exceptionally large and sparse vocabularies,
    thanks to the number of morpheme slots and combinations in a word.
    This complexity, together with a general scarcity of written data,
    poses a challenge to the development of natural language technologies.
    To address this challenge, we offer linguistically-informed approaches for bootstrapping a neural morphological analyzer,
    and demonstrate its application to Kunwinjku, a polysynthetic Australian language.
    We generate data from a finite state transducer to train an encoder-decoder model.
    We improve the model by ``hallucinating'' missing linguistic structure into the training data,
    and by resampling from a Zipf distribution to simulate a more natural distribution of morphemes.
    The best model accounts for all instances of reduplication in the test set and achieves an accuracy of 94.7\% overall, a 10 percentage point improvement over the FST baseline.
    This process demonstrates the feasibility of bootstrapping a neural morph analyzer from minimal resources.
\end{abstract}

\section{Introduction}
Polysynthesis represents the high point of morphological complexity.
For example, in Kunwinjku, a language of northern Australia (ISO gup),
the word \form{ngarriwokyibidbidbuni} contains six morphs:

\begin{exe}
  \ex \label{ex:ngarriwokyibidbidbuni}
  \gll 
  ngarri- wok- yi- bid- bidbu- ni\\
  1pl.excl- word- COM- REDUP- go.up- PI\\
  \glt `We were talking as we climbed up'
\end{exe}

Example (\ref{ex:ngarriwokyibidbidbuni}) illustrates common features of polysynthesis: fusion, incorporation, and reduplication. 
Fusion combines multiple grammatical functions into a single morph, leading to large morph classes,
and challenging the item-and-arrangement leanings of finite state morphology.
Incorporation presents a modelling challenge because rule-based methods are unable to enumerate an open class, and machine learning methods need to learn how to recognize the boundary between contiguous large or open morph classes. 
Reduplication is also a challenge because it copies and prepends a portion of the verb root to itself, requiring a nonlinear or multi-step process.
Tackling these phenomena using finite state transducers (FSTs) involves a combination of technical devices whose details depend on subtleties of the morphological analysis \cite[cf.][]{arppe2017computational}.
There remains a need for more investigation of polysynthetic languages to
deepen our understanding of the interplay between the options on the computational side, and the most parsimonious treatment on the linguistic side.

Morphological complexity leads to data sparsity, as the combinatorial possibilities multiply with each morpheme slot:
most morphologically complex words will be rare.
Furthermore, many morphologically complex languages are also endangered, making it difficult to collect large corpora.
Thus, polysynthetic languages challenge existing ways of building tools and applications for the communities that speak these languages.

In this work we investigate Kunwinjku, spoken by about 2,000 people in West Arnhem in the far north of Australia.
Members of the community have expressed interest in using technology to support language learning and literacy development.
Thus, we face the challenge of developing useful language technologies on top of robust models, with few resources and in a short space of time.
We envisage morphologically-aware technologies including dictionary interfaces, spell checkers, text autocompletion, and tools for language learning \cite[cf.][]{littell2018indigenous}.

This paper is organized as follows.
We begin by reviewing previous work in finite state morphology, low resource morph analysis, neural approaches to morph analysis, and data augmentation for morphological reinflection (Sec.~\ref{sec:background}).
Next, we describe our existing finite state model for Kunwinjku verbs (Sec.~\ref{sec:kmorphfeats}).
In Section~\ref{sec:methods} we present a neural approach which addresses gaps in the previous model,
including the ability to analyze reduplication and to exploit distributional information.
Next we discuss our evaluation metrics and our handling of syncretism and ambiguity (Sec.~\ref{sec:eval}).
Finally, the results are presented in Section~\ref{sec:results},
including a discussion of how well the neural models address the shortcomings of the FST model.

Our contributions include:
(a)~a robust morphological analyzer for verbs in a polysynthetic language;
(b)~a method for augmenting the training data with complex, missing structure; and
(c)~a technique for scoring the likelihood of generated training examples.

\section{Background and Related Work}
\label{sec:background}
Finite state transducers (FSTs) are a popular choice for modelling the morphology of polysynthetic languages.
Several toolkits exist, including XFST, Foma, and HFST \cite{beesley2003finite,hulden2009,linden2013hfst}. 
Each one is an optimized implementation of the finite state calculus \cite{kaplan1994regular}, providing additional support for morphosyntactic and morphophonological processes.
Most recent work on computational modelling of morphologically rich languages is built on the foundation of these tools \cite{arppe2017computational,littell-2018-finite,andriyanets-tyers-2018-prototype,chen2018morphological,cardenas2018morphological}. 
As a case in point, we applied Foma in the analysis of the morphology of Kunwinjku verbs, but ran into difficulties accounting for out-of-vocabulary (OOV) items in open morph classes. We also stopped short of addressing complex features like reduplication and verbal compounding, for technical reasons related to the expressiveness of FSTs \cite[cf.][]{lane2019}. 

Recently, neural models have gained popularity for morphological processing because they address some of the weakness of FSTs:
subword modeling shows an ability to remain robust in the face of out-of-vocabulary items,
and recurrent neural architectures with attention have shown a capacity to learn representations of context which allow the model to incorporate the notion of long-distance dependencies \cite{bahdanau2014neural}.

Neural morphological analyzers can be developed from training data generated by an FST.
These analyzers are more robust, handling variation, out-of-vocabulary morphs, and unseen tag combinations \cite{micher2017improving,moeller2018neural,schwartz2019bootstrapping}. 
They provide 100\% coverage, always providing a ``best guess'' analysis for any surface form. 
Of course, FSTs can be modified to accommodate exceptions and OOV morphs,
but this requires explicit modelling and usually does not achieve the robustness of neural analyzers \cite{schwartz2019bootstrapping}.

\citet{anastasopoulos-neubig-2019-pushing} found that they could augment their training set by hallucinating new stems, increasing accuracy on their test set by 10 percent.
This method involved substituting random characters from the target language's alphabet into the region identified by alignment as the probable root. 
For the sake of cross-lingual generalizability, their method does not consider language-specific structure. 

The task of morphological analysis, mapping an inflected form to its root and grammatical specifications,
is similar to the task of machine transliteration, mapping a sequence of words or characters from source to target language without reordering. 
For example in Kunwinjku, consider the segmentation and gloss of the verb \form{karridjalbebbehni}:

\begin{exe}
  \ex 
  \gll karri- djal- bebbeh- ni\\
  12a- just- DISTR- sit.NP\\
  \glt `Let's just sit down separately' [E.497]
\end{exe}

Since the process of segmenting and glossing the verb does not contain any reorderings, the mapping of surface to glossed forms can be viewed as transliteration.

\section{A Finite State Model of Kunwinjku} \label{sec:kmorphfeats}

Finite state transducers have long been viewed as an ideal framework to model morphology \cite{beesley2003finite}.
They are still a popular choice for low-resource polysynthetic languages \cite[cf.][]{chen2018morphological,lachler2018modeling}. 
Here we summarize some features of Kunwinjku and describe the finite state implementation.

\begin{figure*}[t!]
  \centering
  \resizebox{\textwidth}{!}{%
  \begin{tabular}{|lll|l|l|l|l|l|l|l|l|l|l|ll|}
  \hline
  $-$12   & $-$11     & $-$10    & ($-$9)          & ($-$8)     & ($-$7)    & ($-$6)          & ($-$5) & ($-$4)          & ($-$3)              & ($-$2)            & ($-$1)         & 0         & +1 & +2  \\
  Tense & Subject & Object & Directional & Aspect & Misc1 & Benefactive & Misc2   & GIN & BPIN & NumeroSpatial & Comitative & Verb root & RR & TAM \\ \hline
  \end{tabular}%
  }
  \caption{Verbal affix positions in Kunwinjku. Regions where indices share a cell ([$-$12,$-$10], [+1,+2]) indicate potentially fused segments. Slot indices in parentheses indicate optionality. 
  Adapted from \cite[Fig 8.1]{evans2003pan}.}
  \label{fig:verbal-affix-positions}
  \end{figure*}

\subsection{Features of Kunwinjku}

Kunwinjku is a polysynthetic agglutinating language,
with verbs having up to 15 affix slots (Fig.~\ref{fig:verbal-affix-positions}). 
Morphs combine in a way that is ``almost lego-like'' \cite{evans2003pan,baker2003word}.

\begin{figure}[t!]
\resizebox{\columnwidth}{!}{%

\tikzset{every picture/.style={line width=0.75pt}} 

\begin{tikzpicture}[x=0.75pt,y=0.75pt,yscale=-1,xscale=1]

\draw    (430.5,132) -- (430.5,166) ;
\draw [shift={(430.5,168)}, rotate = 270] [color={rgb, 255:red, 0; green, 0; blue, 0 }  ][line width=0.75]    (10.93,-3.29) .. controls (6.95,-1.4) and (3.31,-0.3) .. (0,0) .. controls (3.31,0.3) and (6.95,1.4) .. (10.93,3.29)   ;

\draw   (259,91) -- (464.5,91) -- (464.5,131) -- (259,131) -- cycle ;
\draw   (262,169) -- (464.5,169) -- (464.5,209) -- (262,209) -- cycle ;
\draw    (430.5,168) -- (430.5,134) ;
\draw [shift={(430.5,132)}, rotate = 450] [color={rgb, 255:red, 0; green, 0; blue, 0 }  ][line width=0.75]    (10.93,-3.29) .. controls (6.95,-1.4) and (3.31,-0.3) .. (0,0) .. controls (3.31,0.3) and (6.95,1.4) .. (10.93,3.29)   ;

\draw    (359.5,210) -- (359.5,244) ;
\draw [shift={(359.5,246)}, rotate = 270] [color={rgb, 255:red, 0; green, 0; blue, 0 }  ][line width=0.75]    (10.93,-3.29) .. controls (6.95,-1.4) and (3.31,-0.3) .. (0,0) .. controls (3.31,0.3) and (6.95,1.4) .. (10.93,3.29)   ;

\draw    (359.5,246) -- (359.5,212) ;
\draw [shift={(359.5,210)}, rotate = 450] [color={rgb, 255:red, 0; green, 0; blue, 0 }  ][line width=0.75]    (10.93,-3.29) .. controls (6.95,-1.4) and (3.31,-0.3) .. (0,0) .. controls (3.31,0.3) and (6.95,1.4) .. (10.93,3.29)   ;

\draw    (359.5,54) -- (359.5,88) ;
\draw [shift={(359.5,90)}, rotate = 270] [color={rgb, 255:red, 0; green, 0; blue, 0 }  ][line width=0.75]    (10.93,-3.29) .. controls (6.95,-1.4) and (3.31,-0.3) .. (0,0) .. controls (3.31,0.3) and (6.95,1.4) .. (10.93,3.29)   ;

\draw    (359.5,90) -- (359.5,56) ;
\draw [shift={(359.5,54)}, rotate = 450] [color={rgb, 255:red, 0; green, 0; blue, 0 }  ][line width=0.75]    (10.93,-3.29) .. controls (6.95,-1.4) and (3.31,-0.3) .. (0,0) .. controls (3.31,0.3) and (6.95,1.4) .. (10.93,3.29)   ;

\draw   (217.5,16) -- (517.5,16) -- (517.5,275) -- (217.5,275) -- cycle ;

\draw (361.75,111) node  [align=left] {morphotactic \\transducer};
\draw (363.25,189) node  [align=left] {morphophonological\\transducers};
\draw (360,257) node  [align=left] {karribimbom};
\draw (354,150) node  [align=left] {karriˆbimˆbuˆ\texttildelow om};
\draw (366,43) node  [align=left] {[V][1pl.incl.3sg.PST][GIN.bim]bu[PP]};
\draw (157,45) node  [align=left] {Analyzed form:};
\draw (150,149) node  [align=left] {Intermediate form:};
\draw (159,256) node  [align=left] {Surface form:};

\end{tikzpicture}
}
\caption{The high-level structure of the Kunwinjku finite state transducer. 
Analyzed forms are mapped to surface forms (and vice versa) through the composition of morphotactic and morphophonological transducers.}
  \label{fig:fst-structure}
\end{figure}
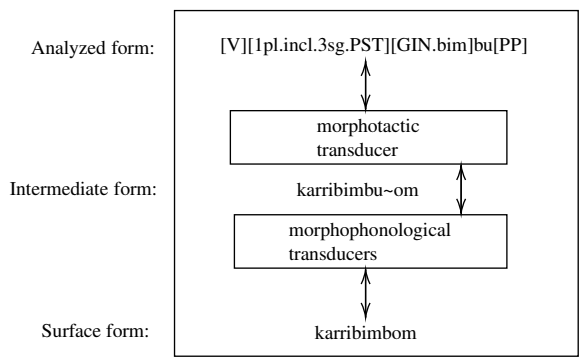

We implement morphotactics and morphophonology as separate stages, following usual practice (Fig.~\ref{fig:fst-structure}).
However, this is not conducive to modelling noun incorporation, valence-altering morphology, fusion, or reduplication,
all typical phenomena in polysynthetic languages.

Kunwinjku has two kinds of noun incorporation.
General incorporable nouns (\morph{GIN}) are a closed class, manifesting a variety of grammatical roles (\ref{ex:ngakakkeleminj}).
Body part incorporable nouns (\morph{BPIN}) are an open class, restricting the scope of the action (\ref{ex:ngabidkeleminj}).

\begin{exe}
  \ex \label{ex:ngakakkeleminj}
  \gll
  nga- kak- keleminj \\
  1m- night- fear.P \\
  \glt `I was afraid at night'
\end{exe}

\begin{exe}
  \ex \label{ex:ngabidkeleminj}
  \gll
  nga- bid- keleminj \\
  1m- hand- fear.P \\
  \glt `I was afraid for my hand' [E.458]
\end{exe}

The open class \morph{BPIN} occupy slot $-3$ and will be adjacent to the verb root whenever slots~$-2$ and~$-1$ are empty, as is common.
With adjacent open class slots, Kunwinjku opens up the possibility of there being \emph{contiguous OOV morphs}.
In Kunwinjku there is no template to help distinguish members of these adjacent classes, thus creating a novel challenge for predicting morph boundaries.

While transitivity of the verb is lexically defined, there are three morph classes which signal valency change:
the benefactive (\morph{BEN}), comitative (\morph{COM}), and reflexive (\morph{RR}). 
More details about the respective function of these morphs is given in \citet{lane2019}, but here it suffices to say their presence in a verb makes resolving valency impossible without wider sentential context.
This impacts the FST modelling, as we are unable to restrict possible illegal analyses on this basis, which results in overgeneration. 

Morphological fusion can lead to a proliferation of morphs and analyses.
In Kunwinjku, there are no fewer than 157 possibilities for the first slot of the verb,
fusing person and number (for both subject and object) along with tense.
We find that this fusion affects decisions around tokenization of the data in preparation for training the seq2seq model (Sec.~\ref{sec:tokscheme}).

Most of the world's languages employ reduplication productively for diverse purposes \cite{rubino2005reduplication}.
It is a common feature of polysynthetic languages in particular.
While modelling reduplication using FSTs is possible, the general consensus is that modelling partially reduplicative processes explode the state space of the model, and are burdensome to develop \cite{culy1985complexity,roark2007computational,dras2012complex}.
For these reasons, the Kunwinjku FST model does not include an implementation of the language's complex reduplication system. 

In Kunwinjku, there are three types of verbal reduplication: iterative, inceptive, and extended. 
Each type of reduplication has 1--3 (CV) templates which can be applied to the verb root to express the semantics associated with each type. 
In Section~\ref{sec:redup} we discuss an approach to ensure that the neural model handles Kunwinjku's complex reduplication system.

\subsection{Evaluating the FST}
\label{sec:evalFST}
We establish a baseline by scoring the FST on a set of $n=304$ inflected verbs.
The data was collected from the Kunwinjku Bible (which targets a modern vernacular),
a language primer \citep{etherington1998kunwinjku},
and a website \cite{bininjkunwokorg}. 
The data was glossed in consultation with language experts.

We define coverage as number of analysed forms, and accuracy as the number of \emph{correctly} analyzed forms, both as a fraction of $n$.
We define precision as the number of correctly analysed forms as a fraction of the number of analysed forms.
We distinguish accuracy and precision because the ability of a model to withhold prediction in case of uncertainty is useful in certain application contexts. 

The results of the evaluation show that while the FST is fairly high-precision, its accuracy is limited by the imperfect coverage of verb stems in the lexicon (Fig.~\ref{fig:fst-eval}). 

\begin{figure}[]
  \centering
  \begin{tabular}{@{}llll@{}}
  \toprule
      & Accuracy & Coverage & Precision \\ \midrule
  FST & 84.4     & 88.5     & 95.4      \\ \bottomrule
  \end{tabular}
  \caption{All-or-nothing accuracy and coverage of the Kunwinjku FST Analyzer on the test set of 304 inflected verbs.}
  \label{fig:fst-eval}
  \end{figure}

The FST relies on a lexicon to provide analyses for inflected forms, and when it comes across OOV morphs, or verb stems modified by processes like reduplication, it fails to return an analysis. 
We sort the coverage issues into classes, and remark that the largest source of error comes from reduplication, followed by variation in tense/aspect/mood (TAM) inflection, OOV stems, OOV incorporated nominals, and exceptions to the d-flapping alternation rule (Fig.~\ref{fig:fst-error}). 
We address each of these problems in the following sections.

\begin{figure}[]
  \centering
  \begin{tabular}{@{}ll@{}}
  \toprule
  Error Class       & \% of Error \\ \midrule
  Reduplication     & 28.9       \\
  TAM Inflection    & 28.5       \\
  OOV root          & 26.3       \\
  OOV inc. nominals & 13.2       \\
  Alternation       & 2.2        \\ \bottomrule
  \end{tabular}
  \caption{Error analysis of \citet{lane2019}'s FST model of Kunwinjku verbs shows 5 classes of error and the percent of the total error attributed to each class.}
  \label{fig:fst-error}
  \end{figure}

\section{Methods}
\label{sec:methods}
In this section we discuss the approach which leverages an incomplete FST to produce a more robust neural morphological analyzer for Kunwinjku. 
Those steps include generating training pairs from an FST, tokenizing the data, resampling from the dataset to simulate distributional signal, hallucinating missing structures into the dataset, and training a neural encoder-decoder model on the resampled data.

\subsection{Data generation from an FST}

Given our low resource setting, training a neural encoder-decoder model like those used in neural machine translation (NMT) is not possible without augmenting what resources we do have. 
Following the established template of recent work on neural morphological analysis for low resource polysynthetic languages \cite{micher2017improving,moeller2018neural,schwartz2019bootstrapping} we use the FST model to generate morphotactically valid pairs of surface and analyzed verbs. 

For the purpose of training the base neural model, we adapted the Foma tool to randomly generate 3,000,000 surface/analysis pairs from the FST (see Fig.~\ref{fig:tokenization} for an example of a tokenized pair).
An automatic process removed duplicates, leaving us with 2,666,243 unique pairs which we partitioned into an .8/.1/.1 train/dev/test split.

In \citet{schwartz2019bootstrapping}'s work on modelling complex nouns in Yupik, they generate a training set which exhaustively pairs every Yupik noun root with every inflectional suffix, regardless of the resulting semantic fidelity. 
In our case, it was not feasible to exhaustively generate the training data, as it would have led to $4.9 \times 10^{12}$ instances (Fig.~\ref{fig:exaustive-combinations}). 
In effect, the training set represents .00004\% of the space over which we seek to generalize.

\begin{figure*}[]
  \resizebox{\textwidth}{!}{
  \begin{tabular}{@{}lllllllllllllllllllllllll@{}}
  \toprule
  TSO &   & DIR &   & ASP &   & MSC1 &   & BEN &   & MSC2 &   & GIN &   & BPIN &   & COM &   & root &   & RR &   & TAM &   & \multicolumn{1}{r}{\textit{Total}} \\ \midrule
  157 & x & 3   & x & 2   & x & 24   & x & 2   & x & 4    & x & 78  & x & 32   & x & 2   & x & 541  & x & 2  & x & 5   & = & $4.9x10^{12}$                          \\ \bottomrule
  \end{tabular}
  }
  \caption{An estimate for all morphotactically valid sequences covered by the Kunwinjku FST}
  \label{fig:exaustive-combinations}
  \end{figure*}

\subsection{Tokenization}
\label{sec:tokscheme}
To prepare the data for training a seq2seq model, we first collect the glossed inflected verb forms, perform tokenization, and organize them into source-target pairs. 

We chose a tokenization scheme which treats graphemes as atomic units.
Morph labels are also treated mostly as atomic units, with the exception being for fused labels which we break into their individual linguistic components (Fig.~\ref{fig:tokenization}). 
For example the pronominal morph in Kunwinjku can simultaneously express both subject and object, as well as tense. 
Consider the pronominal prefix \form{kabenbene-} which we gloss as \morph{3sg.3ua.nonpast} and tokenize as \morph{[ 3sg . 3ua . nonpast ]}. 
Choosing to break up labels in the fused morphological slots prevents an unnecessary proliferation of entries in the target vocabulary, as individual units like \morph{3sg}, \morph{3ua}, and \morph{past} can be shared by multiple pronominals.
Our choice to tokenize the source forms and verb root strings at the grapheme level reflects our desire to loosen the model's vocabulary such that it is equipped to handle variation at the orthographic level, and possible OOV stems. 

\begin{figure*}
  \begin{lstlisting}[frame=single,basicstyle=\small]
  b i k a nj ng u n e ng -> [ 3sg . 3Hsg . PST] [ BPIN ] ng u [ PP ]

  \end{lstlisting}
  \caption{An example of a tokenized source/target training pair, where we treat source graphemes, target labels, fused target label components, and verb root graphemes as atomic units.}
  \label{fig:tokenization}
\end{figure*}

\subsection{Simulating distributional information}

Generating from an FST at random fails to capture valuable information about the distribution of morphs. 
For example in Kunwinjku, body part incorporable nouns (\morph{BPIN}) can occur adjacent to the verb root.
Both categories are open class, meaning that there is a high likelihood in the low-resource setting that either or both are out-of-vocabulary.
How then does the analyzer decide where to place the boundary?
Perhaps the entire sequence is a single out-of-vocabulary root.
Our intuition is that knowing how likely analyzed tag pairs co-occur can help to disambiguate.
Some morph sequences are inevitably more frequent than others, and we would like to represent that information in the training set. 

To this end, we propose a method for simulating distributional information in the training set.
First, we want to score any analyzed form, giving higher scores to forms that contain more likely sequences.
We define $M$ as the sequence of morph tags which make up an analysis, where $m_i$ is the morph tag at index $i$.
The scoring function is defined as follows:

\begin{exe}
  \ex
  \( \textrm{score}(M) =\frac{1}{n}\sum\limits ^{n}_{i} \log P( m_{i} ,m_{i+1}) \)
\end{exe}

The joint probability of adjacent tags is estimated from a corpus of unannotated text, here, selected books from the Kunwinjku Bible.
Everything the existing FST can analyse as a verb is considered to be a verb, and is used to calculate the joint probability table.

The training set is tagged with the FST\footnote{By using an FST with imperfect recall we are not capturing true distributional information; it is simply a heuristic.}, and ranked according to the scoring function.
We split the sorted data into buckets defined by their morphotactic likelihood, and then sample from them according to a Zipf distribution.
The effect is that more probable sequences are more likely to occur in the training data than less likely examples, thus approximating the distribution of morphotactic structure we would expect to see in a natural corpus. 

\subsection{Hallucinating reduplicative structure} 
\label{sec:redup}

One shortcoming of the Kunwinjku FST model is that it does not account for reduplicative structure, due to the complexity of modelling recursive structure in the linear context of finite state machines \cite{culy1985complexity,roark2007computational}. 
As noted previously, reduplication is responsible for 28.9\% of the FST's coverage error when evaluated on the test set of inflected verbs.
If reduplication is not modeled by the FST, then reduplication will also not be represented in the training set generated by that FST. 
We posit that if data hallucination has been shown to improve performance in the language-agnostic setting \cite{anastasopoulos-neubig-2019-pushing,silfverberg-etal-2017-data}, than it is likely that linguistically-informed hallucination can provide a similar reinforcement in Kunwinjku. In line with this, we developed an extension to the data generation process which hallucinates reduplicative structure into a subset of the training data.

Kunwinjku has three main types of partial verbal reduplication signaling iterative, inceptive, and extended meaning.  
Moreover, each type of reduplication can have more than one CV template, depending on which paradigm the verb belongs to.
Figure~\ref{fig:reduplication} documents the three types of reduplication, and serves as the template for the reduplicative structure hallucinator. 

\begin{figure*}[t!]
\centering
\resizebox{\textwidth}{!}{%
\begin{tabular}{|l|lll|l}
\hline
\textbf{Type} & \multicolumn{1}{l|}{\textbf{Pattern(s)}}       & \multicolumn{1}{l|}{\textbf{Unreduplicated Verb}} & \textbf{Reduplicated Verb}                     & \multicolumn{1}{l|}{\textbf{Semantic Effect on Verb (V)}}                          \\ \hline
\multirow{3}{*}{Iterative}     & \multicolumn{1}{l|}{CVC}                       & \multicolumn{1}{l|}{dadjke = cut}                 & dadj-dadjke = cut to pieces                    & \multicolumn{1}{l|}{\multirow{3}{*}{Doing V over and over again}} \\ \cline{2-4}
                                & \multicolumn{1}{l|}{CV(C)CV(h)}                   & \multicolumn{1}{l|}{bongu = drink}               & bongu-bongu = keep drinking                 & \multicolumn{1}{l|}{}                                             \\ \cline{2-4}
                                & \multicolumn{1}{l|}{CVnV(h)}                   & \multicolumn{1}{l|}{re = go}                      & rengeh-re = go repeatedly                      & \multicolumn{1}{l|}{}                                             \\ \hline
\multirow{2}{*}{Inceptive}     & \multicolumn{1}{l|}{\multirow{2}{*}{CV(n)(h)}} & \multicolumn{1}{l|}{yame = spear (sth)}     & yah-yame = try (and fail) to spear (sth) & \multicolumn{1}{l|}{Failed attempt to do V}                       \\ \cline{3-5} 
                                & \multicolumn{1}{l|}{}                          & \multicolumn{1}{l|}{durnde = return}              & durnh-durnde = start returning                 & \multicolumn{1}{l|}{Starting to do V}                             \\ \hline
\multirow{2}{*}{Extended}      & \multicolumn{1}{l|}{CVC(C) \(\Vert\) \_ men}            & \multicolumn{1}{l|}{djordmen = grow}              & djordoh-djordmen = grow all over the place     & \multicolumn{1}{l|}{\multirow{2}{*}{Doing V all over the place}}  \\ \cline{2-4}
                                & \multicolumn{1}{l|}{CVC(C) \(\Vert\) \_ me}                                  & \multicolumn{1}{l|}{wirrkme = scratch}           & wirri-wirrkme = scratch all over               & \multicolumn{1}{l|}{}                                             \\ \hline
\end{tabular}%
}
\caption{Reduplication in Kunwinjku has three forms, and each form has its own CV templates defining how much of the verb is captured and copied.
In the case where we've used the form X \(\Vert\) \_ Y, we mean that pattern X is the reduplicated segment if found in the context of Y.
Figure adapted from \cite{evans2003pan}.}
\label{fig:reduplication}
\end{figure*}

First, the hallucinator module samples n\% of the FST-generated pairs and strips away the affixes to isolate the root.
For each root, one of the three reduplication types (iterative, inceptive, or extended) is selected at random, and the root is matched against the available CV templates.
The longest pattern which matches the root is selected, and the pattern-matching portion of the root is copied and prepended to the root.
Both the surface and analyzed form are updated to reflect the change, and the new training pairs are appended to the original list of FST-generated pairs.

\subsection{Training}

We trained an encoder-decoder model on the dataset of 2,114,710 surface/analyzed form pairs (the Base model).
We then hallucinate reduplication into 8\% of the Base data, and combine that hallucinated data to the base training data set (the Base+halluc[...] models).

The model setup is similar to the one described in \cite{schwartz2019bootstrapping}.
We use MarianNMT: a fast, open-source toolkit which implements neural models for machine translation \cite{junczys2018marian}.
We used a shallow attentional encoder-decoder model \cite{bahdanau2014neural} using the parameters described in \cite{sennrich2016edinburgh}: the encoder and decoder each have 1 hidden layer of size 1024.
We use cross-validation as the validation metric, set dropout to .2 on all RNN inputs, and enable early stopping to avoid overfitting.
We use the same setup and parameters for all NMT models mentioned in this paper. 
A full accounting of the MarianNMT settings used can be seen in the Appendix.

\section{Evaluation of the Neural Models} \label{sec:eval}

We begin by reporting the performance of the neural models in terms of coverage, accuracy, and precision, so that they can be compared with the evaluation of the FST model, described in Section~\ref{sec:evalFST}.
Additionally, we measure the performance of the neural models in terms of precision (P), recall (R), and F1 on the morph level: For each morph tag in the gold target test set, we calculate P, R, and F1, and then calculate the macro-average P, R, and F1 across all tags in the test set (Fig.~\ref{fig:morph-eval}).
This method is more granular than all-or-nothing accuracy over the entire translated sequence, and allows us to get a better picture of how the models are doing on the basis of individual tags.  

We observed an issue with syncretic ambiguity which complicates the evaluation process
(also noted by \citealt{schwartz2019bootstrapping}; \citealt{moeller2018neural}).
For example, the pronominal prefix \form{kabindi-} can be glossed: \morph{[3ua.3ua.nonpast]}, or \morph{[3pl.3ua.nonpast]}, or \morph{[3ua.3pl.nonpast]}, or \morph{[3pl.3pl.nonpast]}.
Here, the pronominal expresses both the subject and object, and is not explicit whether that subject or object is the 3rd person dual or plural, in any of four possible combinations.
The disambiguation cannot be resolved at the level of the isolated verb.

Our initial experiment with the base data set achieved 100\% coverage and 68.3\% accuracy on the test set.
When confronted by the same problem, \citet{moeller2018neural} decided to collapse ambiguous tags into an underspecified meta-tag.
For example, for the Kunwinjku data, we might collapse the four tags above into \morph{[3pl.3pl.nonpast]}.
However, doing so results in a potential loss of information.
Given the wider sentential context, the pronominal could be possibly be disambiguated, so long as the distinction is preserved and all equally-valid analyses are returned.

Further, as \citet{schwartz2019bootstrapping} point out, in the Yupik language it is possible for this ambiguity to exist across other categories which are not easily collapsed.
In Kunwinjku, an example of this would be the pronominals \morph{[1sg.2.past]} and \morph{[3sg.past]} which differ in terms of number and valency, and yet share the same null surface form.
Their differences are such that they can not be easily collapsed into a single meta-tag. 
Therefore we do not penalize the model for producing any variation of equally valid analyses given the surface form, and for each model we adjust the evaluation for syncretism in a post-processing step.

\section{Results and Discussion} \label{sec:results}

All of the neural models outperform the FST in terms of accuracy and coverage (Fig.~\ref{fig:acc-eval}). 
However, the FST is more precise, and this may be useful in certain application contexts. 
The best model is Base+halluc+resample, which improves on the FST by 10.3 percentage points. 
On the morph-level, we see that the neural models containing the hallucinated reduplication data outperform  the base neural model (Fig.~\ref{fig:morph-eval}).  

\begin{figure}[h]
  \resizebox{\columnwidth}{!}{%
  \begin{tabular}{llll}
                & Acc & Cov & Precision \\
  FST           & 84.4     & 88.5 & \textbf{95.4}     \\
  Base        & 89.1     & \textbf{100}  & 89.1   \\
  Base+halluc & 93.7     & \textbf{100}  & 93.7  \\
  Base+halluc+resample & \textbf{94.7} & \textbf{100} & 94.7 \\
\end{tabular}
  }
  \caption{All-or-nothing accuracy and coverage of the three morphological analyzer models}
  \label{fig:acc-eval}
  \end{figure}

\begin{figure}[h]
  \resizebox{\columnwidth}{!}{%
  \begin{tabular}{llll}
                & Precision    & Recall    & F1   \\
  Base        & 88.8 & 89.9 & 89.0 \\
  Base+halluc & 91.6 & 92.6 & 91.8 \\
  Base+halluc+resample & \textbf{93.7} & \textbf{93.6} & \textbf{93.4} \\
  \end{tabular}
  }
  \caption{Morph-level performance of shallow neural sequence models. Macro P/R/F1 across all morph tags.}
  \label{fig:morph-eval}
  \end{figure}

We posited that the difficulties encountered by the FST model---namely reduplication, out-of-vocabulary items, and spelling variation---could be at least partially addressed by training a neural model on character and tag sequences, and hallucinating instances of reduplication into the training set.
For the most part, this held true, as we see gains across all error classes (cf.~Sec.~\ref{sec:evalFST}).
Here we report performance with respect to the three largest error classes: reduplication, OOV verbs, and OOV nouns. 

\subsection{Reduplication}

As expected, neither the FST nor the Base neural model succeeds in recognizing reduplication.
It would be impossible, as the \morph{REDUP} tag does not appear in either of their vocabularies. 

The Base+halluc model's performance gain over the Base model can be accounted for entirely by the fact that it achieved 100\% recall of reduplicative structure.
Precision, on the other hand was 57.9\%.
Looking at the errors, we find that the imprecise predictions were all applied to instances about which the system was already wrong in previous models, meaning that the impact of reduplicative hallucination between models was only positive.
In the Base+halluc+resample model, recall of reduplicative structure was also 100\%, and precision increased slightly to 58.8\%.

\subsection{Discovering New Lexical Items}

The neural models correctly identify some unseen verb stems, but still show room for improvement.
We observe a tendency across all neural models to predict verb stems which have been seen in training, and which are also a substring of the observed unknown root.
For example, the training set does not contain any verbs with the root \form{dolkka}, but it shows up 3 times in the test set.
The analyses of all \form{dolkka}-rooted verbs were the same in both the Base+halluc and Base+halluc+resample models: they propose \form{ka}, a known root from the training set, and presume \form{dolk-} to be an incorporable noun\footnote{Possibly by virtue of its orthographic proximity to \form{bolk-}, a common general incorporable noun which means ``land.''}.
Figure~\ref{fig:root-guesses} shows a sample of OOV verb stems and nouns from the test set.
In the unseen verbs table, this behavior of preferring previously observed verb stems is the cause of error in every case.  

\begin{figure}[]
  \resizebox{\columnwidth}{!}{%

  \begin{tabular}{llc}
  \textbf{Unseen Verbs}     & \textbf{Base+halluc+resample}     & \textbf{\checkmark /\xmark} \\
  \textbf{wobekka}ng            & [GIN]bekka             & \xmark   \\
  nga\textbf{kohbanjm}inj       & [GIN][REDUP]me      & \xmark     \\
  nga\textbf{rrukkendi}         & dukkendi                   & \checkmark  \\
  ka\textbf{menyime}            & [GIN]yime            & \xmark        \\
  yimalng\textbf{darrkiddi}     & darrke[PERSIST] & \xmark       \\
  ngam\textbf{dolkka}ng         & [DIR][GIN]ka       & \xmark        \\
  \textbf{dolkka}ng             & [GIN]ka              & \xmark        \\
  ka\textbf{rrukmirri}          & dukmirri                   & \checkmark   \\
  ngurrimirnde\textbf{mornname}rren& mornname & \checkmark \\\\
  \textbf{Unseen GIN/BPIN/ASP}     & \textbf{Base+halluc+resample}     & \textbf{\checkmark /\xmark} \\
  kan\textbf{njilng}marnbom     & [GIN]                & \xmark  \\
  yiben\textbf{kange}marnbom & [REDUP] & \xmark \\
  kan\textbf{kange}murrngrayekwong & [GIN] & \xmark \\
  kankange\textbf{murrng}rayekwong & [BPIN] & \checkmark \\
  kankangemurrng\textbf{rayek}wong & [REDUP] & \xmark \\
  kan\textbf{kange}marnbom  & [REDUP] & \xmark \\
  ngarri\textbf{bangme}marnbuyi & [BPIN] & \xmark \\
  yi\textbf{malng}darrkiddi     & [GIN][REDUP] & \xmark        \\

  \end{tabular}
  }
  \caption{Column 1 shows the list of verbs and nouns (in bold) which are are unseen in the FST lexicon.
  Column 2 is the Base neural model's prediction covering the character sequence corresponding to the unseen item.
  Column 3 indicates whether the neural model's analysis of the morph is correct. }
  \label{fig:root-guesses}
  \end{figure}
  
Further difficulty comes in distinguishing between general (\morph{GIN}) and body-part (\morph{BPIN}) incorporated noun classes.
The low rate of success in positing unknown incorporated nouns is, in large part, attributed to the fact that the large \morph{GIN} and open \morph{BPIN} classes often occur adjacent to each other and to the root.
The neural model has difficulty making useful predictions when multiple morphs in this region are previously unobserved.

Overall, the Base+halluc+resample model correctly posited 33\% of unseen stems, and 12.5\% of unseen nouns from the FST error analyses.

\subsection{Impact of distributional information}
This technique to approximate distributional information led to a small improvement in overall accuracy, and in tag-level P/R/F1.
We had expected that this information might help the neural models learn something about the relative frequencies of \morph{GIN}s or \morph{BPIN}s, which could help make decisions about how to draw the boundary between unseen stems and unseen incorporated nominals. 
Instead, we saw distributive information helped to disambiguate the boundaries between morph classes with fewer members.

One representative example is the case of \form{yikimang}, whose root is \form{kimang}.
Before resample, the neural models interpret the \form{yi-} as the comitative prefix \form{yi-}, and injects a spurious \morph{COM} tag into the analysis.
After resample, it correctly omits the \morph{COM} tag, interpreting \form{yi-} as the 2nd person singular pronominal.
In the unfiltered FST-generated training data, \morph{COM} occurs in 53\% of instances.
In the resampled data, it occurs in 22\% of instances.
When all morph labels are equally likely to occur, the model is just as likely to predict any morph label compatible with the character sequence.
Resampling the training data according to a more realistic distribution leads to stronger morph transition priors, which tip the scale in favor of the analysis with a more likely tag sequence.

\section{Conclusion} \label{sec:conclusion}

We have shown that complex features of polysynthetic morphology, such as reduplication and distributional morphotactic information, can be simulated in the dataset and used to train a robust neural morphological analyzer for a polysynthetic language.
In particular, we showed that a robust neural model can be bootstrapped in a relatively short space of time from an incomplete FST.

This work represents a successful first iteration of a process whereby the morphological model can be continually improved.
Indeed, the concept of \emph{bootstrapping} a model implies an iterative development story where much of the scaffolding used in early efforts will eventually fall away.
For example, once the bootstrapped model has been used to tag verbs containing reduplication, we can confirm the model's high-confidence predictions and retrain.
In this second iteration, we may find that we no longer need to hallucinate reduplication because it is sufficiently represented in the new training set.
Similarly, once we have applied the complete neural model to a corpus of natural text, we will no longer need to approximate distributional information.
For researchers developing robust morphological analyzers for low resource, morphologically complex languages, this work represents a template of model development which is well-suited for the context.

Producing a viable morphological analyzer is the first step towards building improved dictionary search interfaces, spell-checking tools, and computer-assisted language learning applications for communities who speak low-resource languages.
The pattern of training robust systems on data that has been augmented by the knowledge captured in symbolic systems could be applied to areas outside of morphological analysis, and is a promising avenue of future exploration.

\section*{Acknowledgments}
We are grateful for the support of the Warddeken Rangers of West Arnhem. 
This work was covered by a research permit from the Northern Land Council,
and was sponsored by the Australian government through a PhD scholarship,
and grants from the Australian Research Council and the Indigenous Language and Arts Program.
We are grateful to four anonymous reviewers for their feedback on an earlier version of this paper.

\vfil\pagebreak

\bibliographystyle{acl_natbib}
\bibliography{acl2020}

\vfil\pagebreak

\appendix
\section*{Appendix}
\label{sec:appendix}
We provide the MarianNMT configuration settings used for all neural models in this work.

\begin{verbatim}
  --type amun 
  --dim-vocabs 600 500 
  --mini-batch-fit -w 3500 
  --layer-normalization 
  --dropout-rnn 0.2 
  --dropout-src 0.1 
  --dropout-trg 0.1 
  --early-stopping 5 
  --valid-freq 10000 
  --save-freq 10000 
  --disp-freq 1000 
  --valid-metrics cross-entropy 
  --overwrite 
  --keep-best 
  --seed 1111 
  --exponential-smoothing 
  --normalize=1 
  --beam-size=12 
  --quiet-translation  
\end{verbatim}

\end{document}